\documentclass[a4paper, 10pt, conference]{ieeeconf}      

\IEEEoverridecommandlockouts                              

\usepackage{amsmath} 
\usepackage{amssymb}  

\usepackage{multirow}
\usepackage{tabularx}

\usepackage{todonotes}
\usepackage{tikz}

\let\labelindent\relax
\usepackage{enumitem}

\usepackage{subcaption}

\usepackage[left=1.59cm,right=1.59cm,top=1.91cm,bottom=2.54cm]{geometry}    

\usepackage{balance}

\usepackage{fancyhdr}

\title{\LARGE \bf
Matching Input and Output Devices and Physical Disabilities for Human-Robot Workstations
}

\author{Carlo Weidemann$^{1,\dag}$, Nils Mandischer$^{2,\dag}$, and Burkhard Corves$^{1}$
\thanks{*This work was co-funded by the German Federal Ministry of Labour and Social Affairs from the ``Ausgleichsfond'' as part of the project ``Inklusion und Integration durch Cobots auf dem ersten Arbeitsmarkt'' (AGF.00.00009.22) and by RWTH Innovation GmbH in ``Innovation Sprint''.}
\thanks{\dag Both authors contributed equally to this research.}
\thanks{$^{1}$C. Weidemann and B. Corves are with the Institute of Mechanism Theory, Machine Dynamics and Robotics, RWTH Aachen University, 52062 Aachen, Germany
        {\tt\small weidemann@igmr.rwth-aachen.de}}%
\thanks{$^{2}$N. Mandischer is with the Chair of Mechatronics at University of Augsburg, 86159 Augsburg, Germany
        {\tt\small nils.mandischer@uni-a.de}}%
}

\fancypagestyle{specialfooter}{%
  \fancyhf{}
  
  \fancyfoot[L]{\footnotesize This work was accepted by the International Conference on Systems, Man, and Cybernetics, Kuching, Malaysia, 2024.\linebreak
  © 2024 IEEE. Personal use of this material is permitted.  Permission from IEEE must be obtained for all other uses, in any current or future media, including reprinting/republishing this material for advertising or promotional purposes, creating new collective works, for resale or redistribution to servers or lists, or reuse of any copyrighted component of this work in other works.}
}

\begin{document}

\maketitle
\thispagestyle{empty}
\pagestyle{empty}

\begin{abstract}
    As labor shortage is rising at an alarming rate, it is imperative to enable all people to work, particularly people with disabilities and elderly people. Robots are often used as universal tool to assist people with disabilities. However, for such human-robot workstations universal design fails. We mitigate the challenges of selecting an individualized set of input and output devices by matching devices required by the work process and individual disabilities adhering to the Convention on the Rights of Persons with Disabilities passed by the United Nations. The objective is to facilitate economically viable workstations with just the required devices, hence, lowering overall cost of corporate inclusion and during redesign of workplaces. Our work focuses on developing an efficient approach to filter input and output devices based on a person's disabilities, resulting in a tailored list of usable devices. The methodology enables an automated assessment of devices compatible with specific disabilities defined in International Classification of Functioning, Disability and Health. In a mock-up, we showcase the synthesis of input and output devices from disabilities, thereby providing a practical tool for selecting devices for individuals with disabilities.
\end{abstract}

\setcounter{footnote}{2}
\thispagestyle{specialfooter}
\section{Introduction}
Many people with severe physical disabilities cannot work in assembly and quality assurance without technical aids. In many countries, companies are obliged by law to include people with disabilities (PwD). Meanwhile, the labor shortage is ever-present. Due to over-aging demographics and the trend towards less immigration, the gap between open positions and skilled laborers is growing and there is no turning point in sight. However, enabling skilled people to participate who would otherwise not be able to work due to congenital (PwD) or acquired (elderly, accident victims) disabilities, can become this exact turning point.

In ongoing projects, we deal with the partially automated design and programming of human-robot assembly workstations for PwD\footnote{In the following the term PwD will be used as collective term for all kind of disabilities, including congenital, acquired, and temporary disabilities.}. At the workstations, assembly and quality inspection tasks are to be carried out, while the robot takes over the work steps that the PwD cannot carry out themselves. The support is designed on an individual level, tailored to the individual PwD's needs. Based on the nature of the interaction, only physical interaction and assistance is considered. In this work, we propose a method for the selection of suitable input and output (I/O) devices that are both necessary for the process and that can be operated by the specific PwD. By embedding the method in an interactable mock-up, a lenient selection of I/O devices is possible, even for designers who are usually not confronted with design for PwD.

First, we give an introduction to the International Classification of Functioning, Disability and Health and how PwD are currently accommodated on the labor market -- with a focus on human-robot workstations (Section~\ref{sec:state_of_the_art}). Second, we propose a novel method for device selection (Section~\ref{sec:method}). Therefore, we discuss the structure of a human-robot workstation and according I/O devices required for controlling the majority of assembly processes. We then list suitable devices for PwD and match them with diverse types of disabilities, categorized into limb functions and cognitive capabilities. The method is embedded into an interactable mock-up (Section~\ref{sec:application}), that we showcase in two exemplary scenarios from recently completed projects.

Summarizing, our main contributions are:
\begin{itemize}
    \item Structuring of human-robot workstation and categorization of I/O devices relative to their usability within a work process.
    \item Comprehensive analysis of the degree of disabilities of diverse partial disabilities with impact on the selection of I/O devices.
    \item Implementation of an interactable mock-up for I/O device selection in human-robot workstations~\cite{Zenodo2024}.
\end{itemize}

\section{Related Work}
\label{sec:state_of_the_art}
The World Health Organisation (WHO)~\cite{WHO_2001_BOOK} defines ``[d]isability [as] an umbrella term for impairments, activity limitations and participation limitations. It denotes the negative aspects of the interaction between a person’s health condition(s) and that individual’s contextual factors (environmental and personal factors)''. In the following, we discuss the classification of such disabilities (\mbox{Section~\ref{ssec:icf}}) and how PwD are currently accommodated on the labor market with focus on human-robot workstations (Section~\ref{ssec:accomodation}).

\subsection{International Classification of Functioning, Disability and Health}
\label{ssec:icf}
The International Classification of Functioning, Disability and Health (ICF)~\cite{WHO_2001_BOOK} provided by the WHO serves as a comprehensive framework for organizing information on functioning and disability, offering a standardized language and conceptual basis for defining and measuring health-related states. It aims to facilitate communication across various sectors, enable data comparison across different contexts and time frames, and provides a systematic coding scheme for health information systems. As an accepted United Nations social classification, the ICF aligns with international human rights mandates and national legislation, making it instrumental in monitoring aspects of disability rights. Guided by principles of universality, parity, neutrality, and environmental influence, the ICF recognizes functioning and disability as multi-dimensional concepts influenced by interactions between health conditions, environmental factors, and personal factors. It adopts a biopsychosocial model, integrating social and medical perspectives, and emphasizes the importance of considering multiple dimensions and perspectives in assessing disability. Ultimately, the ICF offers a dynamic understanding of functioning, enabling descriptive profiling rather than binary classifications, thereby, supporting varied measurement and policy applications tailored to specific purposes and contexts, which makes it imperative to base the following methods on the ICF standard.

\subsection{Accommodation for People with Disabilities and its Barriers in Workplace Inclusion}
\label{ssec:accomodation}
Ward and Baker~\cite{WB2005} identify three accommodation strategies, namely assistive technologies, systemic personal change, and universal design, whereas the former two aim at inclusion into an existing work environment. Nevala et al.~\cite{NPK2014} find ``moderate evidence'' that workplace accommodation promotes employment of PwD and may reduce costs in the long run. Syma~\cite{Sym2018} evaluates that technical assistance may also support invisible disabilities as defined by the Invisible Disabilities Association~\cite{ida}, e.g., depression or ADHD. However, Rojas et al.~\cite{RBM2024} and Vujica Herzog, Buchmeister, and Harih~\cite{VBH2019} emphasize that there is a distinct lack in expert knowledge on ergonomic workplace design accommodating PwD. Karbasi, Ansari, and Schlund~\cite{KAS24} give indicators to analyze the inclusivity of work systems. They point out that evidence is missing on qualitative assessment of inclusivity. Hagner, Dague, and Phillips~\cite{HDP2014} indicate that workplace inclusion does not necessarily face a barrier only in technical and ergonomic design, but also in social inclusion, e.g., through lack in on-site socializing or company events.

One way to include PwD into work processes is the use of human-robot workstations. Herzog and Mina~\cite{HM2024} show that PwD initially already accept to work with a collaborative robot and that acceptance improves with interaction time. Arboleda et al.~\cite{APL2020} analyze, i.a., the way how PwD like to control a robot, indicating that control strategies shall comply with individual needs and operate on varying and selectable levels of autonomy. Mark et al.~\cite{MRB2021} analyze functional requirements for assistive workstations. Their workstation design embeds a collaborative robot for screwing tasks controlled by gestures. We designed human-robot workstations for manufacturing~\cite{GLK2022} and quality assessment~\cite{WHF2022}. All workstations were either equipped with universal I/O devices or highly individualized based on expert knowledge (and user feedback). Both options raised the overall cost of the workstation for the including company -- through higher salaries or parts costs. Padkapayeva et al.~\cite{PPY2017} analyze literature on workplaces provided for inclusion of PwD. They observe a broad variety of options for workplace designs, but also a distinct lack in effectiveness of inclusion and cost-effectiveness as selection and design criteria.

This is where our method aims to improve on the state of the art: by simplifying the design process in a human- and inclusion-centric way, hence, lowering overall costs of human-robot workstations for PwD.

\section{Design Method for Automated Selection of Input and Output Devices}
\label{sec:method}
In the following, we present the methodology and preparatory work for the automated selection of I/O devices. The method is oriented along VDI~2221~\cite{vdi2221}, which is commonly used for designing new workstations, and which allows later integration of our method into the framework of the VDI standard. First, we analyze the structure of a human-robot workstation and indicate implications on its design process (Section~\ref{ssec:structure}). Second, we indicate which devices are required in a work process (Section~\ref{ssec:required_devices}). While most processes may be controlled using just push buttons, we discuss why more elaborate control inputs need to be considered. From the identified device types, concrete devices are derived and a morphological box is crafted (Section~\ref{ssec:morph_box}). Last, we evaluate devices on their usability given specific degrees of (partial) disabilities (Section~\ref{ssec:matching}). To this end, we introduce relevant categories of disabilities based on the Degree of Disability Table defined by the German Federal Ministry of Justice~\cite{FMJ_2009}, which are further broken down into degrees of disability to ease the application of our method.

\begin{figure*}[t!]
        \centering
        \includegraphics[width=.95\textwidth]{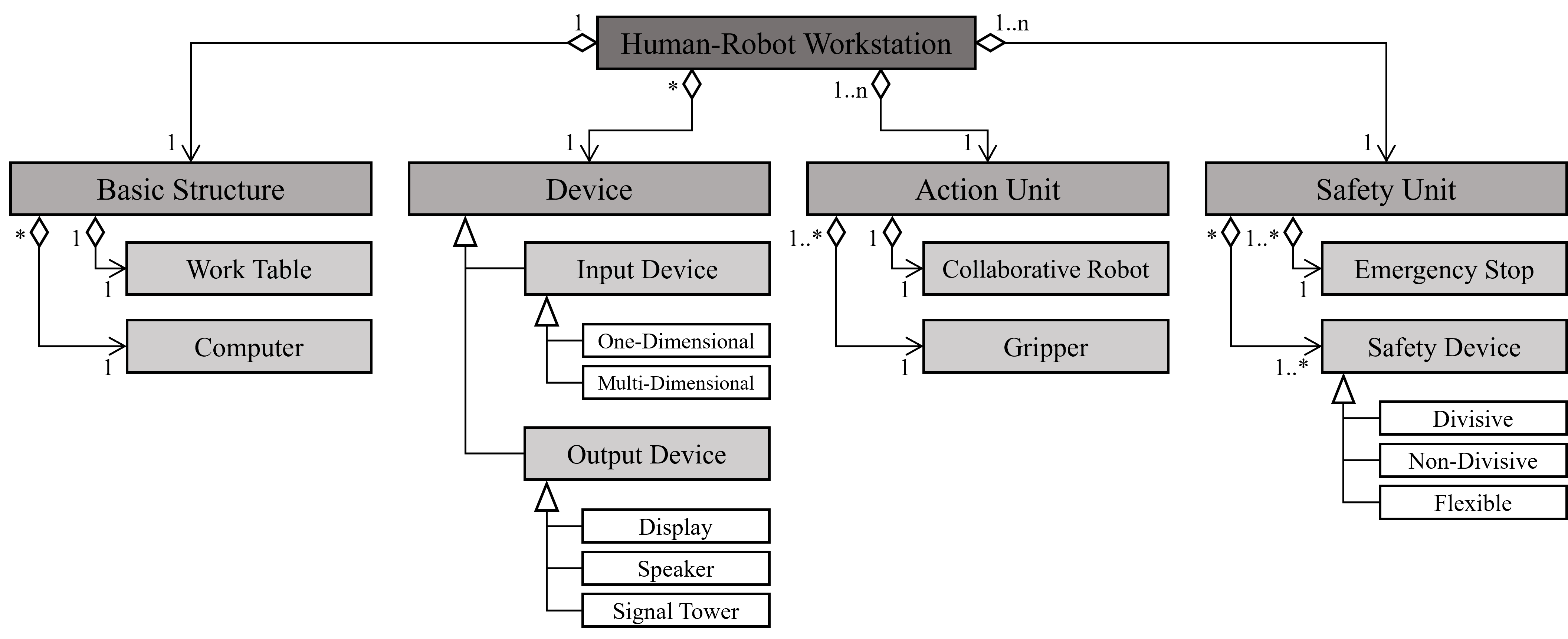}
        \caption{UML diagram of the human-robot workstation, indicating system and subsystem structure and relation between components.}
        \label{fig:structure_plan}
\end{figure*}

\subsection{Structure of Human-Robot Assembly Workstations}
\label{ssec:structure}
We divide a human-robot workstation into four main systems: \emph{Basic Structure}, \emph{Device}, \emph{Action Unit}, and \emph{Safety Unit}. Most systems have subsystems and each subsystem inhibits the workstation components. Figure~\ref{fig:structure_plan} depicts the structure of a typical workstation and the semantic relation of its components. The \emph{Basic Structure} defines the framework for the workstation, establishing hardware interfaces and the workstation computer. The work table serves as the foundation for mounting additional components. The workstation computer serves as higher level control unit (in addition to the robot control). For example, it controls the display or playback of work instructions on the output devices. The \emph{Action Unit} aggregates all acting components of the human-robot workstation, i.e., the collaborative robot and its gripper(s). Each robot typically requires its own safety components. These are aggregated in the \emph{Safety Unit}. For each \emph{Action Unit} there must exist a \emph{Safety Unit}, indicated by the shared parameter~$n$. The safety unit is subject to \mbox{DIN EN ISO 13850}~\cite{ISO_13850} and \mbox{DIN EN ISO 10218-1}~\cite{ISO_10218-1}. Finally, the workstation uses a certain number of input and output \emph{Devices}, which establish and control the interaction between robot and human. There may exist processes which do not use any devices, hence, the missing minimum of aggregated \emph{Devices} in Figure~\ref{fig:structure_plan}. However, in \mbox{Section~\ref{ssec:required_devices}}, we will discuss why processes usually require devices, particularly, if PwD are involved.

While \emph{Basic Structure}, \emph{Action Unit}, and \emph{Safety Unit} must be aggregated to establish the human-robot workstation, the selection of \emph{Devices} is dependent on process (Section~\ref{ssec:required_devices}) and person (Sections~\ref{ssec:morph_box} and \ref{ssec:matching}), which complicates the design process, particularly, for non-experts. Hence, one of the main aspects of our method is to evolve from an expert design process to a documentation procedure applicable and available for all willing to include PwD.

\subsection{Required Devices Based on Work Processes' Structure}
\label{ssec:required_devices}
In order to implement a robot-assisted work process, it is imperative to determine the necessary devices to control the robot (input) or receive information from the workstation (output). In many work processes, the robot must interact with the items and assemblies (parts) by moving or exerting force on them, which demands precise control, e.g., during press-fitting of solenoid coils~\cite{GLK2022}. Manual guidance of the robot does not meet this accuracy requirement, is not economic (in terms of time and cost), and may not be usable at all based on diverse disabilities. Thus, the robot motion usually must be implemented as a position control system, executing pre-programmed poses in a predefined sequence. The actual trajectories are not relevant for the selection of devices. The devices merely function as control unit to choose from a variety of robot poses or behaviors, or to confirm the fulfillment of preceding process steps.

Processes are categorized in sequential and flexible processes. In \textbf{sequential} processes, the robot poses are always executed in the same order and parts are initially in the same position and orientation. The processing of the parts is done at a predetermined position by the human-robot team. Hence, control of sequential processes through I/O devices is simple and streamlined. On the contrary, \textbf{flexible} processes allow multiple options and, therefore, require the opportunity to select from different choices. Examples for such processes are in case the assembly sequence is irrelevant or in case a component must be chosen from a box of identical or different parts. While the selection from choices may be implemented through specific sequences of button presses using a single button (e.g., press once for left, twice for right, etc.), this procedure may take inadmissibly long and raises the probability of errors. Latter is particularly serious when there are multiple similar sequences. Instead, a joystick -- or similar multi-dimensional input device -- is adequate in such cases. 

In addition to the input devices relative to the sequencing of process steps, usable output devices need to be considered. These devices convey system-generated information to users. Common output devices include speakers for auditory work instructions or warning signals, monitors for displaying work instructions, or signal towers for indicating the process status (e.g., \emph{Idle} or \emph{Error}). Each device plays a critical role in facilitating effective communication, enhancing overall usability and user experience.

\begin{table*}[t!]
    \def\tabwidth{.16\textwidth}
    \def\picwidth{.14\textwidth}
    \caption{Morphological box of I/O devices considered in this work, ordered into three categories.}
    \centering
    \begin{tabular}{p{\tabwidth}|p{\tabwidth} p{\tabwidth} p{\tabwidth}|p{\tabwidth}}
        \textbf{One-Dimensional Input} & \multicolumn{3}{c|}{\textbf{Multi-Dimensional Input}} & \centering\arraybackslash\textbf{Output}\\
        \hline
        \centering\arraybackslash\includegraphics[width=\picwidth]{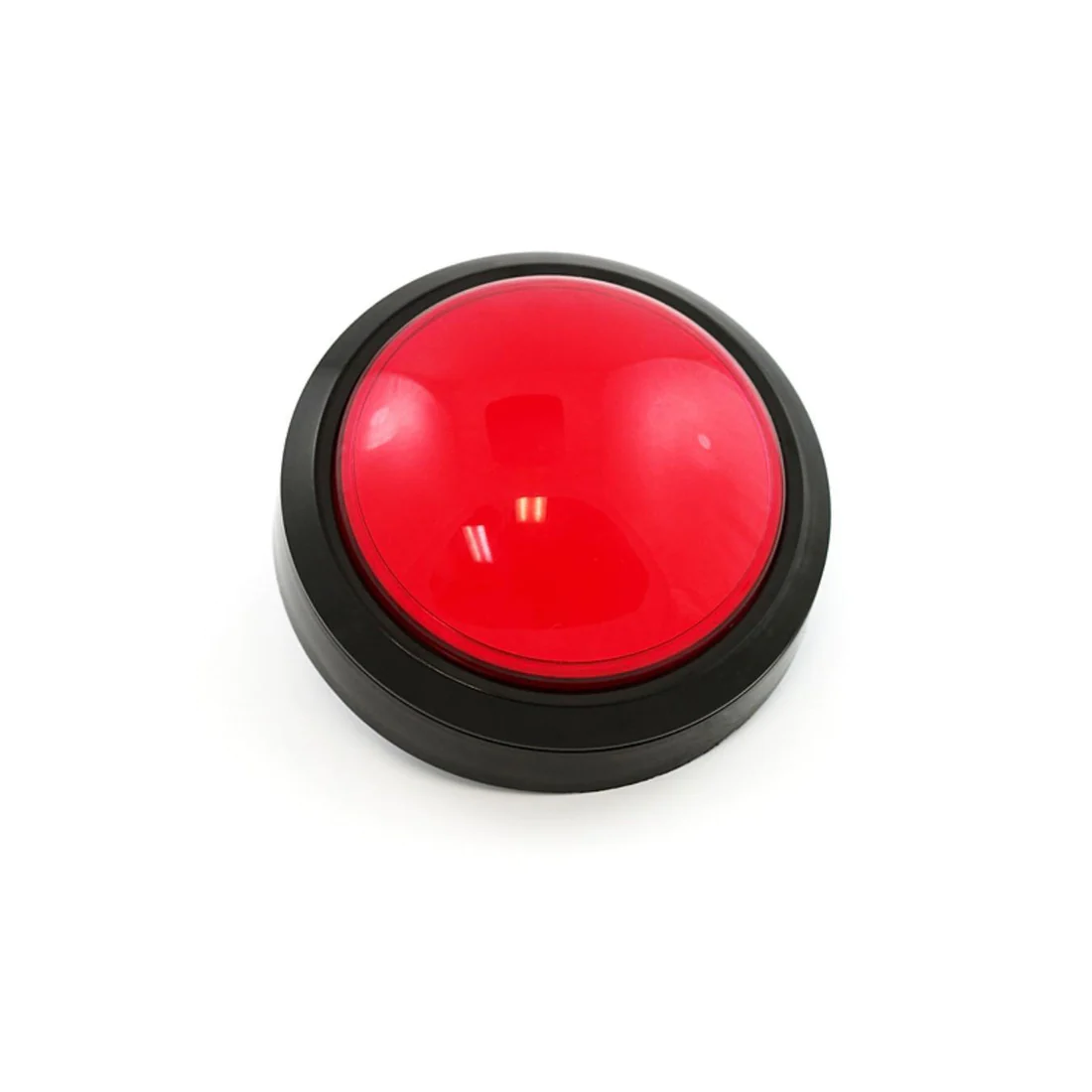} &
        \centering\arraybackslash\includegraphics[width=\picwidth]{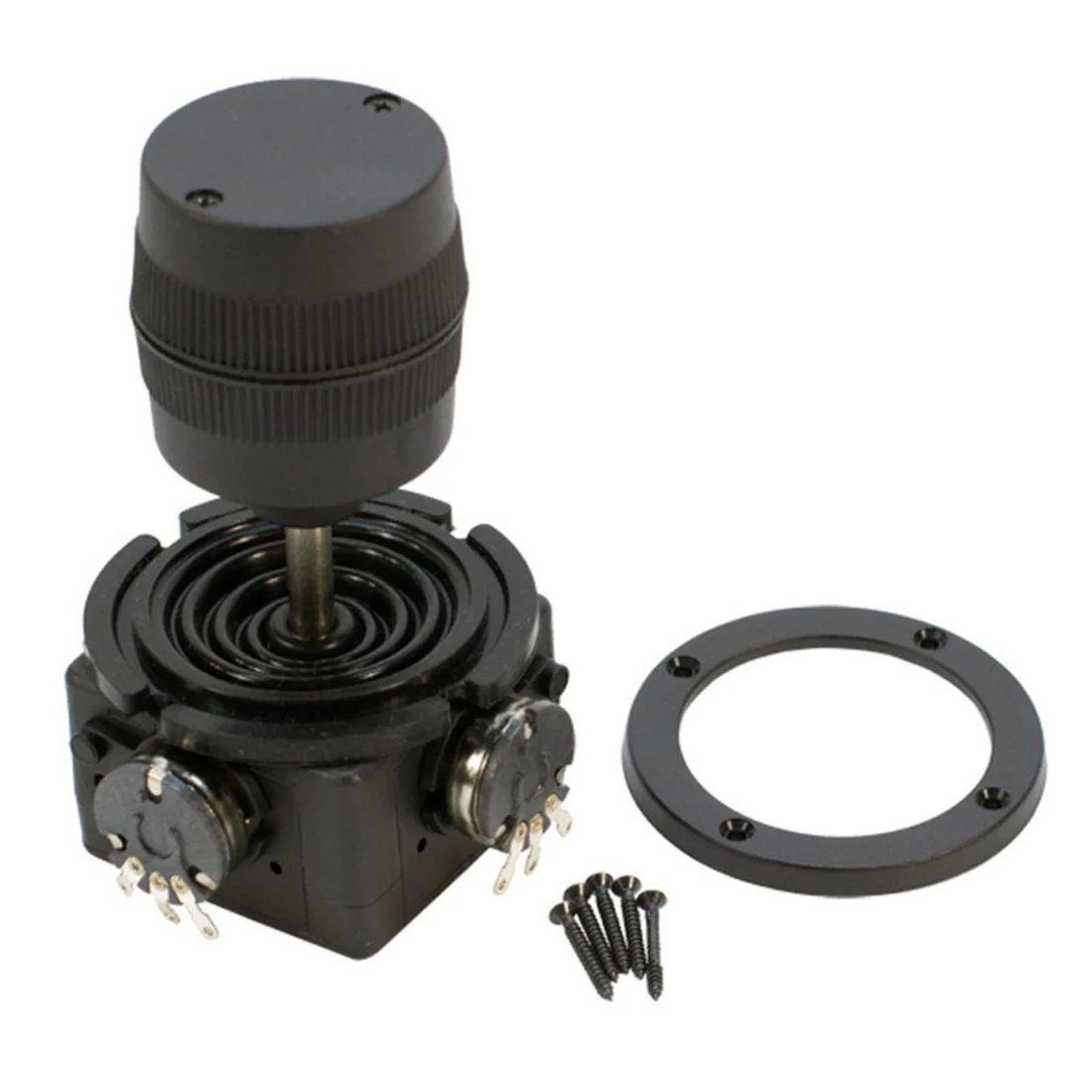} & 
        \centering\arraybackslash\includegraphics[width=\picwidth]{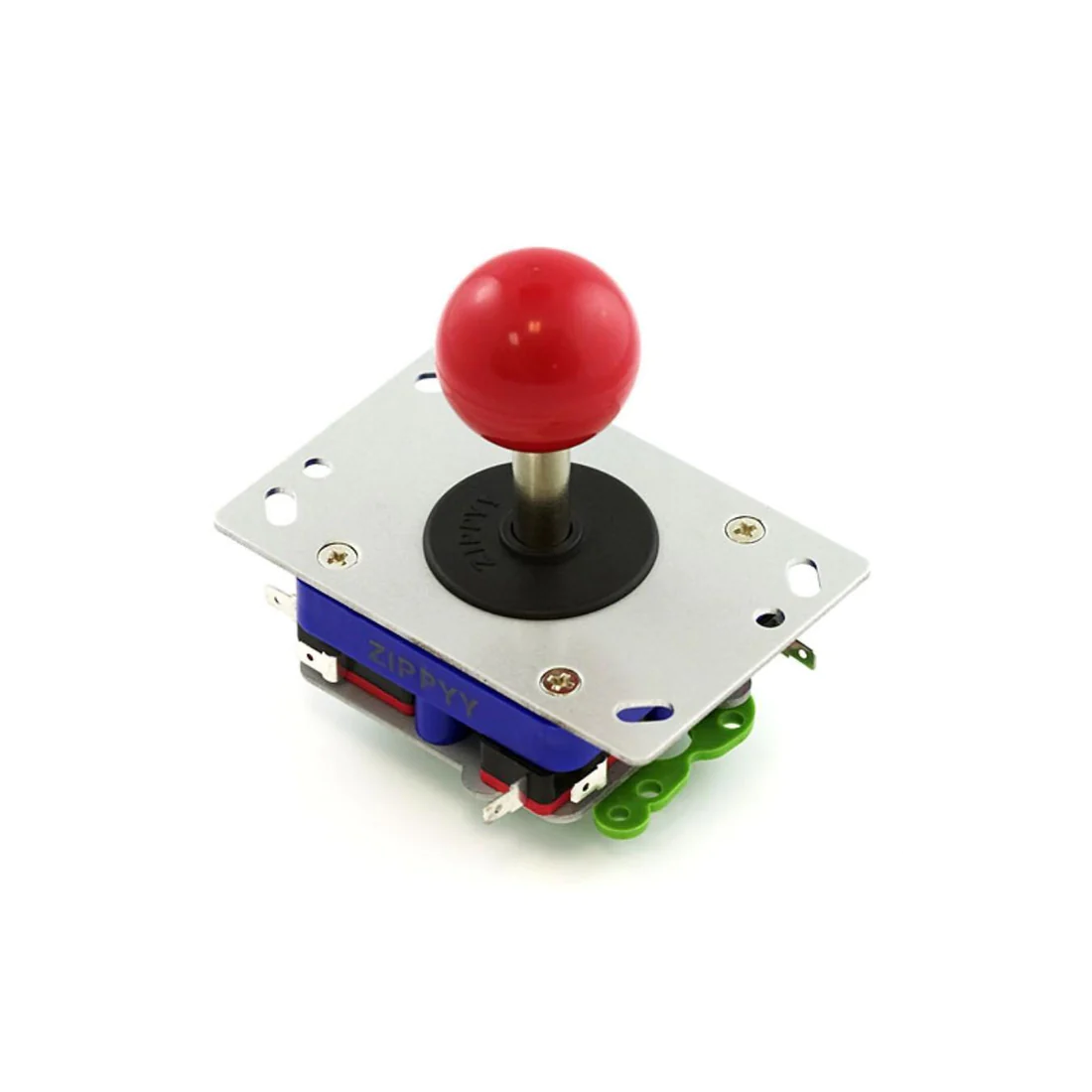} & 
        \centering\arraybackslash\includegraphics[width=\picwidth]{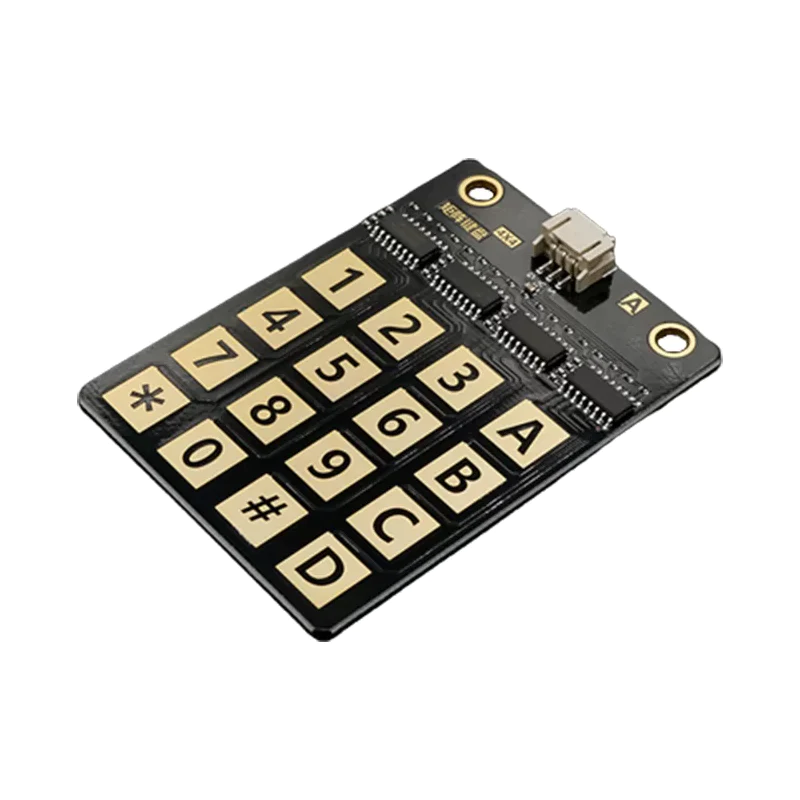}  &
        \centering\arraybackslash\includegraphics[width=\picwidth]{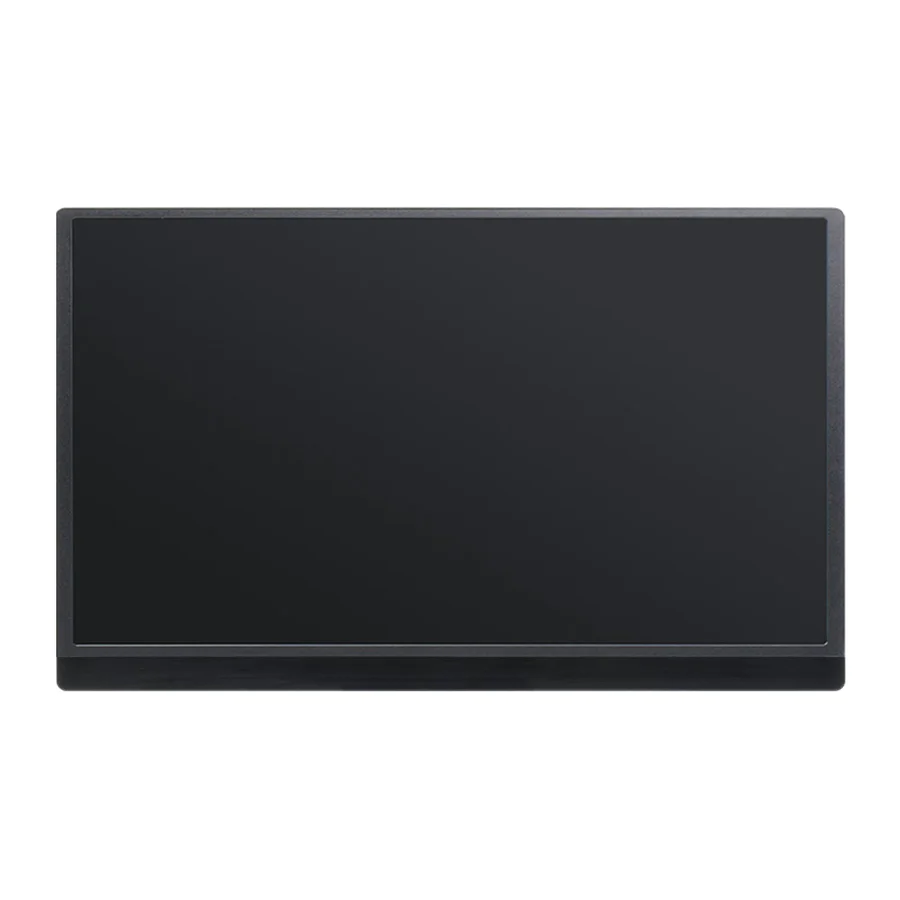}\\
        \centering\arraybackslash hand button~\cite{robotshop} &
        \centering\arraybackslash analog joystick~\cite{robotshop} &
        \centering\arraybackslash digital joystick~\cite{robotshop} &
        \centering\arraybackslash keyboard~\cite{robotshop} &
        \centering\arraybackslash display~\cite{robotshop}\\
        \centering\arraybackslash\includegraphics[width=\picwidth]{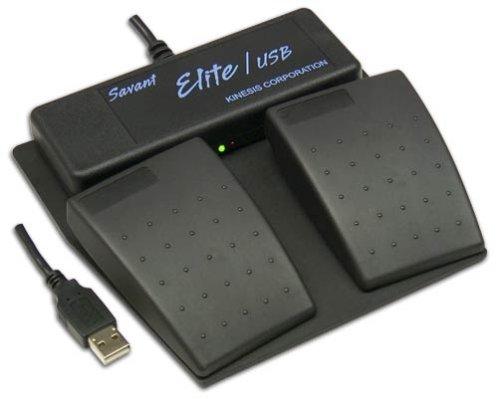} &
        \centering\arraybackslash\includegraphics[width=\picwidth]{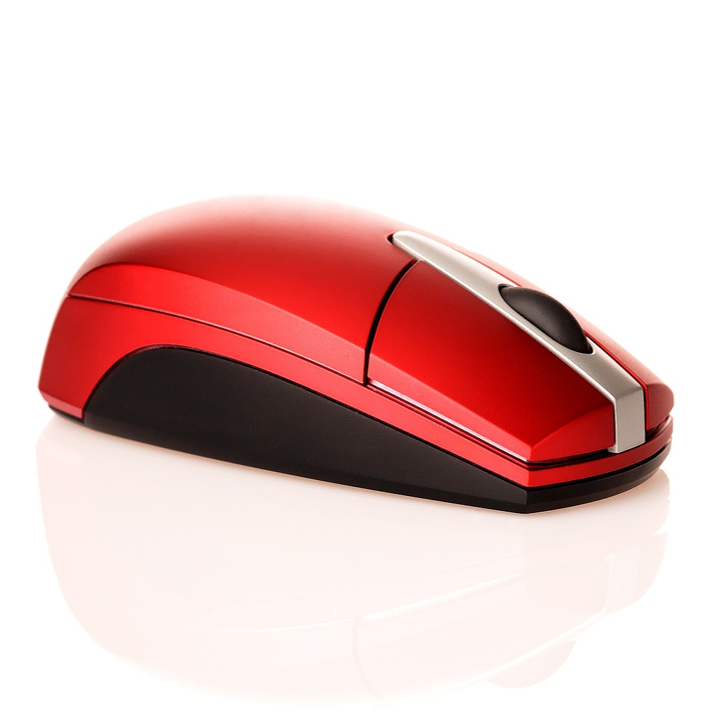} &
        \centering\arraybackslash\includegraphics[width=\picwidth]{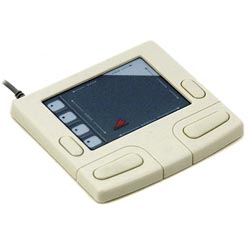} &
        \centering\arraybackslash\includegraphics[width=\picwidth]{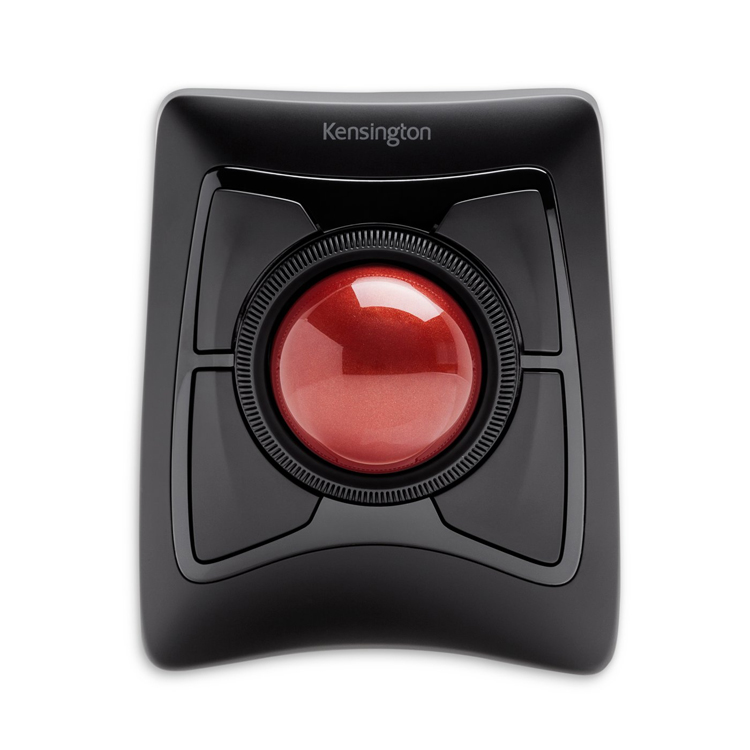} &
        \centering\arraybackslash\includegraphics[width=\picwidth]{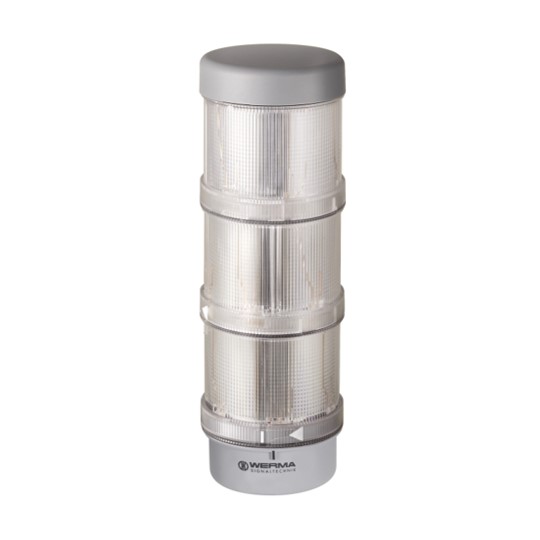}\\
        \centering\arraybackslash foot button~\cite{hws} &
        \centering\arraybackslash mouse &
        \centering\arraybackslash touchpad~\cite{hws} &
        \centering\arraybackslash trackball mouse~\cite{hws} &
        \centering\arraybackslash signal tower~\cite{werma}\\
        ~ & \centering\arraybackslash\includegraphics[width=\picwidth]{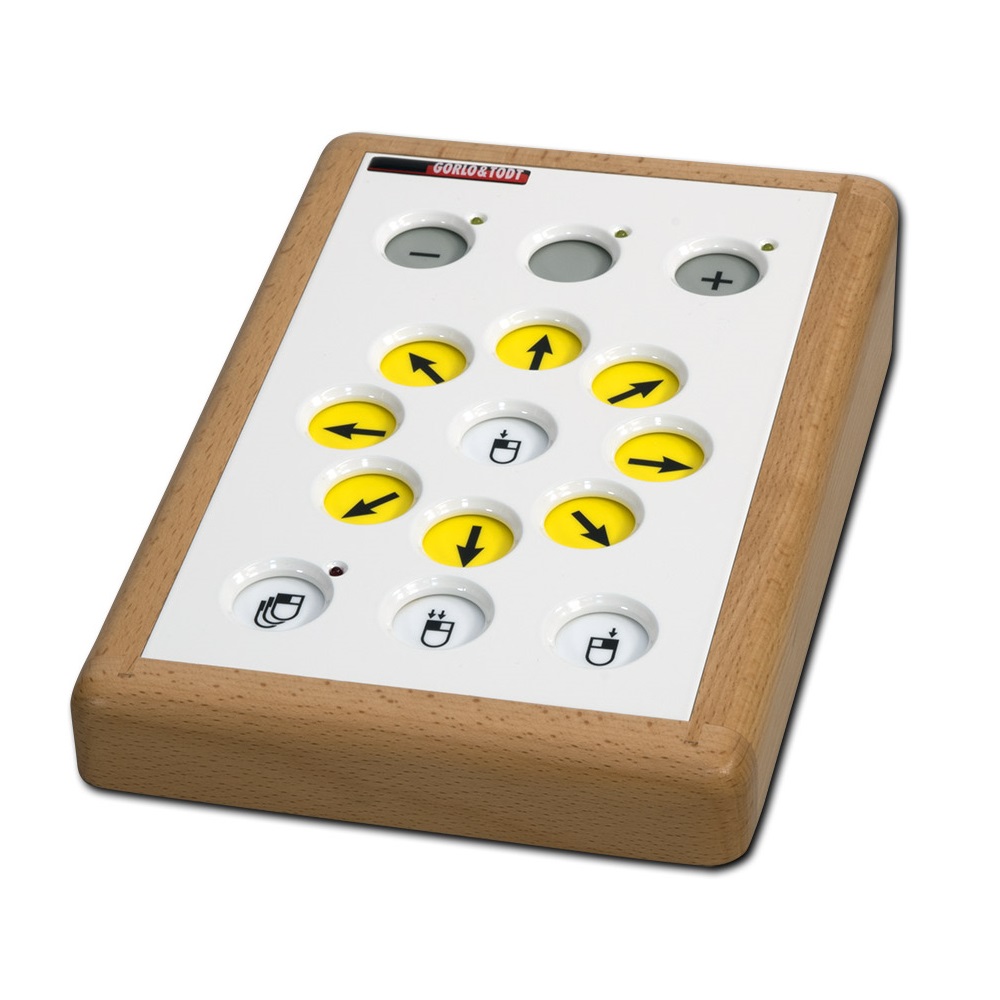} &
        \centering\arraybackslash\includegraphics[width=\picwidth]{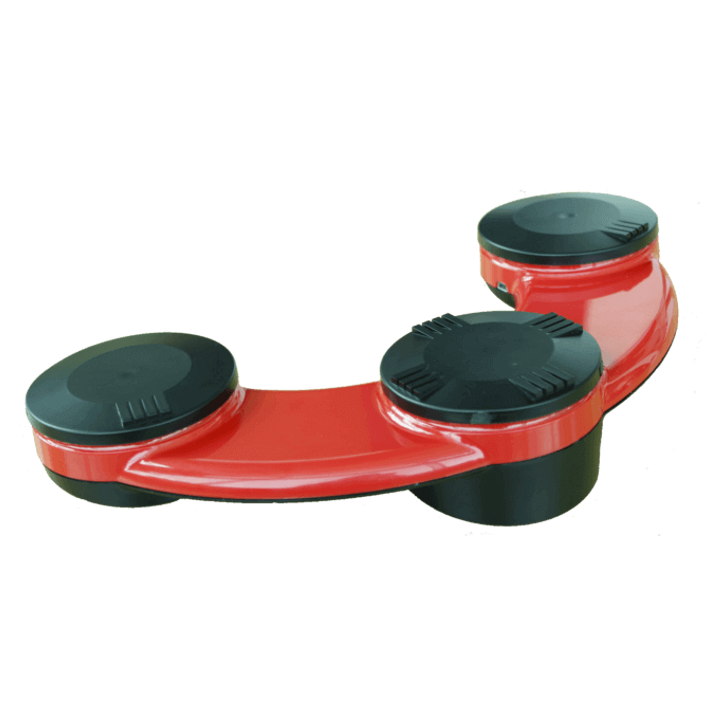} &
        \centering\arraybackslash\includegraphics[width=\picwidth]{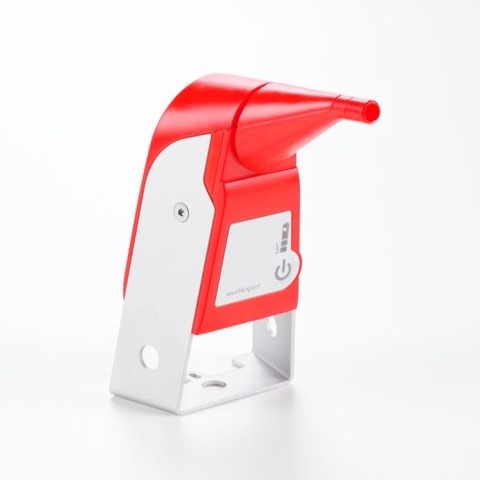} &
        \centering\arraybackslash\includegraphics[width=\picwidth]{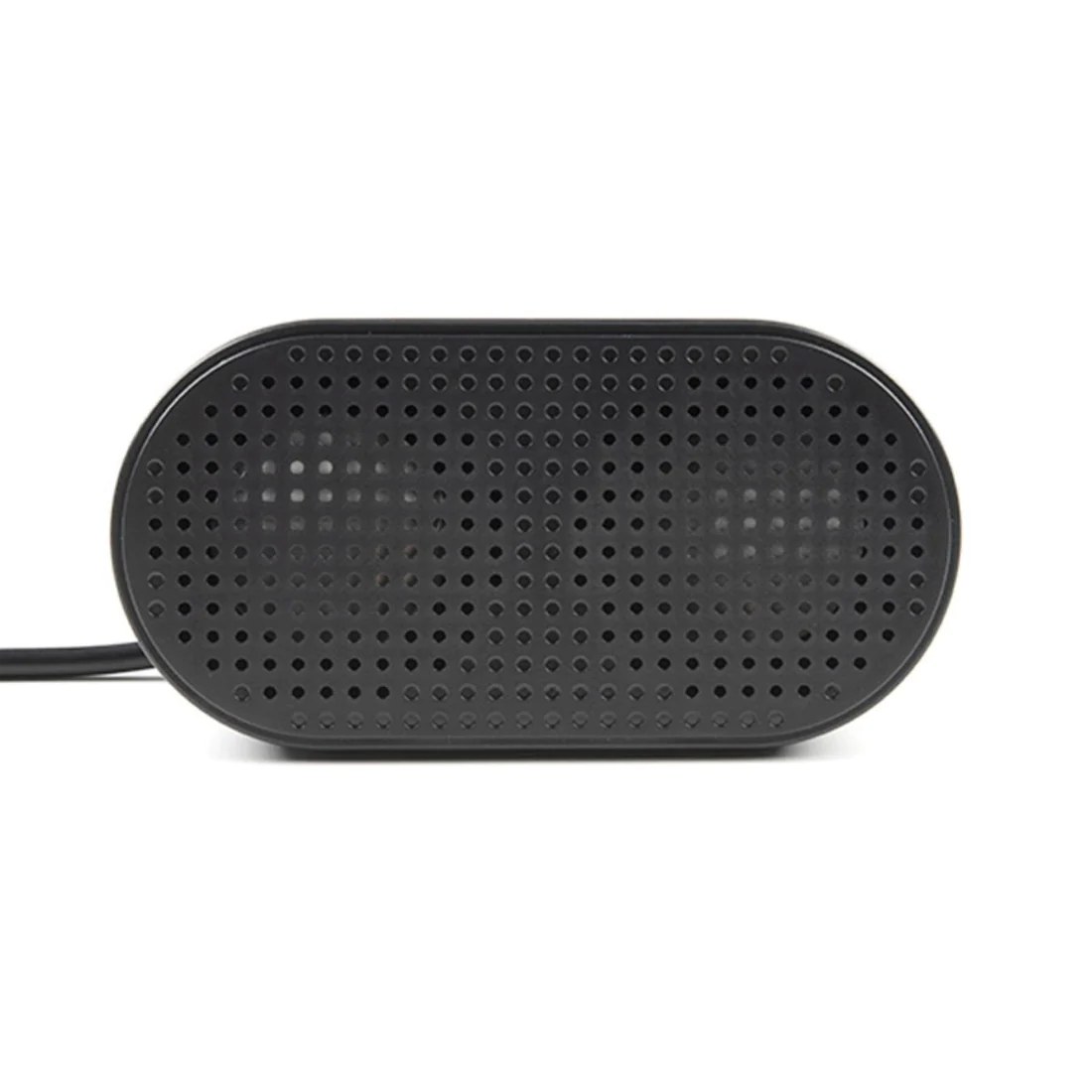}\\
        ~ & \centering\arraybackslash key mouse~\cite{meier} &
        \centering\arraybackslash foot mouse~\cite{dencept} &
        \centering\arraybackslash mouth mouse~\cite{meier} &
        \centering\arraybackslash speaker~\cite{robotshop}
    \end{tabular}
    \label{tab:control_elements}
\end{table*}

\subsection{Suitable Devices for People with Disabilities}
\label{ssec:morph_box}

The I/O devices serve as an interface for interaction between the robot and the human. Those must be adapted to PwD in order to ensure operation of the workstation and an ergonomic working environment. Table~\ref{tab:control_elements} displays the I/O devices that are suitable for PwD and which are to be considered in this work in form of a morphological box.

\subsubsection{Input Devices}
The selection of a suitable button depends on the disability\footnote{In the following, we will use the term ``disability'' as a singular disability or as an aggregation of disabilities (disability complex). In other works, the term ``limitation'' is used for a singular disability.} of the person. There is a variety of buttons in various sizes, shapes, colors, and functional principles (e.g., push buttons, or capacitive buttons), allowing flexible and individual design. If operation by hand is not possible, a foot button may be used. We refrain from considering speech input (which may function as one- or multi-dimensional input), as it is (a)~complex to implement (particularly for the non-expert), (b)~subject to uncertainty\footnote{A button has 100\% accuracy, while the accuracy of speech processing is dependent on the used method.}, and (c) potentially not available in a manufacturing environment with noise. We also do not consider a button to be pressed with other body parts than the arm or foot due to ergonomic issues.

A combination of multiple buttons makes it possible to control complex processes (e.g., writing with a keyboard). A digital joystick essentially consists of four buttons (discrete signal). On the contrary, analog joysticks are suitable for controlling the speed of the robot (e.g., for rotating a component during a visual quality check) due to their continuous signal. A computer mouse is an equivalent to the analog joystick. There exist many kind of mouse specific for PwD that allow the usage with different hand postures. Other alternatives include touchpads for PwD with a limited radius of action and trackballs for PwD with coordination difficulties and muscle weakness. The keyboard mouse is suitable for people with limited fine motor skills. A combination of foot buttons and a foot trackpad may be used in case the arms are restricted, and mouth joysticks if the legs are restricted as well. Note that all multi-dimensional input devices may also be used as one-dimensional input devices.

\subsubsection{Output Devices}
The output devices convey signals from the workstation to the person. This includes the robot status, such as error messages, and work instructions. Typical output devices for the visual transfer of information are signal towers and displays. The signal tower usually consists of three or four lights in the colors red, yellow, green and blue. It is particularly suitable for displaying the status of the system (e.g., red for error or green for running; compare~\cite{WHF2022}). Note that the meaningfulness of the light tower's information depends on the knowledge of the user on the color coding. A display offers a high degree of freedom in design. In addition, the contrast, brightness and size of elements on the display can be adjusted for individual needs, e.g., in case of myopia. Displays are particularly suitable for showing easy-to-assess work instructions in form of images or videos. Speakers make it possible to play warning tones or auditive work instructions. In noisy working environments, headphones are an alternative. These are not depicted in the morphological box as they are essentially just another manifestation of speakers. When selecting output devices for PwD, the two-senses principle should be used. This states that information should always be communicated using at least two senses~\cite{ISO_18040-2}. In the context of collaborative workplaces, only hearing and sight are applicable. Although tactile feedback is an alternative, the precision of the information is limited. In addition, visual and auditory feedback devices are readily available and established in industry.

\subsection{Matching Disabilities and Devices}
\label{ssec:matching}
Until now, we have analyzed which type of devices are mandated by the workstation and work process, and have indicated which variations of devices are suitable for PwD. In the following, we analyze which specific combination of disabilities prevents the use of latter devices with the aim to give recommendations for the design of the workstation, specific for person and process. Note that this method is also applicable in absence of disabilities. To this end, the relevant disabilities and the degrees of disability must be determined. Cognitive and psychological disabilities are not considered because the robot can only provide physical assistance. According to ICF, most physical disabilities can be categorized as performance impairments (e.g., high blood pressure), sensory impairments (e.g., blindness), motor impairments (e.g., Parkinson's syndrome or paralysis), and missing or malformed limbs (e.g., amputations or dysmelia)~\cite{WHO_2001_BOOK}.

\subsubsection{Disability Categories}
The German Degree of Disability Table\footnote{translated from German: ``Grad der Behinderung-Tabelle''.}~\cite{FMJ_2009} lists the most common disabilities. The table is based on the bio-psychosocial model of the ICF. From this list, we select disabilities that influence the use of I/O devices. Physical disabilities affecting joints and muscles are limb loss or malformations, and restricted mobility due to stiffening of joints. These are denoted by the categories \emph{Amputation \& dysmelia} and \emph{Mobility of limbs}. Nerve damage results in \emph{Paralysis}, \emph{Disturbance of movement patterns}, or \emph{Sensitivity to pressure}. Perceptual disabilities affect \emph{Vision} and \emph{Hearing}. Table~\ref{tab:disabilities} lists these seven categories differentiated by the corresponding body part or sense (\emph{Arm}, \emph{Leg} and \emph{Perception}).

\begin{table*}[t!]
    \def\tabwidth{.09\textwidth}
    \newcolumntype{C}[1]{>{\centering\arraybackslash}m{#1}}
    \newcommand{\splitcell}[2]{\centering\begin{tabular}{m{.04\textwidth} m{.04\textwidth}}#1 & #2\end{tabular}}
    \newcommand{\splitcellmore}[2]{\centering\begin{tabular}{m{.06\textwidth} m{.06\textwidth}}#1 & #2\end{tabular}}
    \centering
    \caption{Requirement matrix indicating the maximal degree of disability feasible for operation of a device. Higher degrees prevent operation of the device. Empty cells indicate no correlation between the specific disability category and the device. Values in columns are split into arm (left) and leg (right). In case of no split, values are the same for both limbs.}
    \begin{tabular}{r|C{.1\textwidth}|C{\tabwidth}|C{\tabwidth}|C{.13\textwidth}|C{\tabwidth}|C{\tabwidth}|C{\tabwidth}}
        \multirow{3}{*}{\textbf{Device}} & \multicolumn{5}{c|}{\textbf{Arm or Leg}} & \multicolumn{2}{c}{\textbf{Perception}} \\ \cline{2-8} 
        & Amputation \& Dysmelia & Mobility of Limbs & Paralysis & Disturbance of Movement Patterns & Sensitivity to Pressure & Vision & Hearing \\ \hline
        Hand button & \splitcell{3}{-} & \splitcell{1}{-} & \splitcell{1}{-} & - & \splitcell{0}{-} & - & - \\ \hline
        Foot button & \splitcell{-}{0} & \splitcell{-}{1} & \splitcell{-}{0} & \splitcellmore{-}{1} & \splitcell{-}{0} & - & - \\ \hline
        Analog joystick & \splitcell{2}{-} & \splitcell{0}{-} & \splitcell{0}{-} & \splitcellmore{0}{-} & \splitcell{0}{-} & 1 & - \\ \hline
        Digital joystick & \splitcell{2}{-} & \splitcell{0}{-} & \splitcell{0}{-} & \splitcellmore{1}{-} & \splitcell{0}{-} & - & - \\ \hline
        Keyboard & \splitcell{1}{-} & \splitcell{0}{-} & \splitcell{0}{-} & \splitcellmore{0}{-} & \splitcell{0}{-} & 0 & - \\ \hline
        Mouse & \splitcell{0}{-} & \splitcell{0}{-} & \splitcell{0}{-} & \splitcellmore{0}{-} & \splitcell{0}{-} & 1 & - \\ \hline
        Touchpad & \splitcell{2}{-} & \splitcell{1}{-} & \splitcell{0}{-} & \splitcellmore{0}{-} & \splitcell{0}{-} & 1 & - \\ \hline
        Trackball mouse & \splitcell{2}{-} & \splitcell{1}{-} & \splitcell{0}{-} & \splitcellmore{1}{-} & \splitcell{0}{-} & - & - \\ \hline
        Key mouse & \splitcell{1}{-} & \splitcell{1}{-} & \splitcell{0}{-} & \splitcellmore{1}{-} & \splitcell{0}{-} & 0 & - \\ \hline
        Foot mouse & \splitcell{-}{0} & \splitcell{-}{0} & \splitcell{-}{0} & \splitcellmore{-}{0} & \splitcell{-}{0} & 0 & - \\ \hline
        Mouth mouse & - & - & - & - & - & 0 & - \\ \hline
        Display & - & - & - & - & - & 0 & - \\ \hline
        Signal tower & - & - & - & - & - & 0 & - \\ \hline
        Speaker & - & - & - & - & - & - & 0 \\
    \end{tabular}
    \label{tab:requirement_matrix}
\end{table*}

\subsubsection{Degrees of Disability}
The degree of disability indicates the severity of impact on the usage of a device. In general, the more sever and complex the disability, the less devices are usable. We define the degrees as broad as possible yet as fine as required to allow better generalization of the method. The granularity of the degrees is selected such that each change in value impacts at least one device. Additionally, the degrees must be comprehensible to non-medical personnel to ease application. The degree of disability is analyzed for each device within the morphological box (see Table~\ref{tab:control_elements}) yielding the necessary degrees of disability in Table~\ref{tab:disabilities}. Table~\ref{tab:requirement_matrix} depicts the requirement matrix, which indicates the required degree of disability for a device. For example, the operation of a hand button requires movement of the hand, arm, or arm stump along an axis, typically with minimal force. The degree at which operation becomes infeasible for each category is established as follows:
\begin{description}
    \item[Amputation \& dysmelia] Hand buttons are typically operated using fingers or the entire hand. However, operation using parts of the forearm is possible and feasible. Therefore, the degree at which operation becomes infeasible is when parts of the upper arm are missing, which would otherwise lead to non-ergonomic bent postures.
    \item[Mobility of limbs] Operation requires only small movements. Therefore, immobility of the complete arm (e.g., due to stiffening of all arm joints) prevents operation.
    \item[Paralysis] Operation is obscured. However, as long as some movement in the arm remains (i.e. only paralysis of the fingers and/or hand), operation is feasible. Operation becomes impossible only when paralysis affects the entire arm.
    \item[Disturbance of movement patterns] Minor disturbances do not impede operation, because rough motor skills are sufficient for operation.
    \item[Sensitivity to pressure] Even mild impairment, precludes operation of buttons.
    \item[Vision] Does not impact operation of buttons.
    \item[Hearing] Does not impact operation of buttons.
\end{description}

As we can now indicate which devices are required and which are usable for a specific PwD, we can derive a method for the automated selection of devices. The method is implemented in a mock-up application, which will be introduced in Section~\ref{sec:application}.

\begin{table}[t!]
    \centering
    \caption{Disability categories and degrees of disability ordered into \emph{arm}, \emph{leg}, and \emph{perception}. Degrees are annotated with raising impact on device operation.}
    \begin{tabular}{p{0.2cm}|p{3cm}|p{3.6cm}|p{0.1cm}}
    	~ &
    	\centering\arraybackslash{\textbf{Disability category}} & \multicolumn{2}{c}{\textbf{Degree of disability}} \\
        \hline
        \multirow{17}{*}{\rotatebox[origin=c]{90}{\textbf{Arm}}} & Amputation \& Dysmelia & no limitation & 0 \\
        \cline{3-4}
        ~ & ~ & from 4 fingers & 1 \\
        \cline{3-4}
        ~ & ~ & all fingers & 2 \\
        \cline{3-4}
        ~ & ~ & from the hand & 3 \\
        \cline{3-4}
        ~ & ~ & from parts of the upper arm & 4 \\
        \cline{2-4}
        ~ & Mobility of Limbs & no limitation & 0 \\
        \cline{3-4}
        ~ & ~ & limited mobility of the hand & 1 \\
        \cline{3-4}
        ~ & ~ & limited mobility of the arm & 2 \\
        \cline{2-4}
        ~ & Paralysis & no limitation & 0 \\
        \cline{3-4}
        ~ & ~ & paralysis of the hand & 1 \\
        \cline{3-4}
        ~ & ~ & paralysis of the arm & 2 \\
        \cline{2-4}
        ~ & Disturbance of & no disturbance & 0 \\
        \cline{3-4}
        ~ & Movement Patterns & mild disturbance & 1 \\
        \cline{3-4}
        ~ & ~ & severe disturbance & 2 \\
        \cline{2-4}
        ~ & Sensitivity to Pressure & no limitation & 0 \\
        \cline{3-4}
        ~ & ~ & moderate & 1 \\
        \cline{3-4}
        ~ & ~ & severe & 2 \\
        \hline
        \multirow{15}{*}{\rotatebox[origin=c]{90}{\textbf{Leg}}} & Amputation \& Dysmelia & no limitation & 0 \\
        \cline{3-4}
        ~ & ~ & foot & 1 \\
        \cline{3-4}
        ~ & ~ & from parts of the lower leg & 2 \\
        \cline{2-4}
        ~ & Mobility of Limbs & no limitation & 0 \\
        \cline{3-4}
        ~ & ~ & slightly limited mobility & 1 \\
        \cline{3-4}
        ~ & ~ & severely limited mobility & 2 \\
        \cline{2-4}
        ~ & Paralysis & no limitation & 0 \\
        \cline{3-4}
        ~ & ~ & paralysis of the foot & 1 \\
        \cline{3-4}
        ~ & ~ & paralysis of the leg & 2 \\
        \cline{2-4}
        ~ & Disturbance of & no disturbance & 0 \\
        \cline{3-4}
        ~ & Movement Patterns & mild disturbance & 1 \\
        \cline{3-4}
        ~ & ~ & severe disturbance & 2 \\
        \cline{2-4}
        ~ & Sensitivity to Pressure & no limitation & 0 \\
        \cline{3-4}
        ~ & ~ & moderate & 1 \\
        \cline{3-4}
        ~ & ~ & severe & 2 \\
        \hline
        \multirow{6}{*}{\rotatebox[origin=c]{90}{\textbf{Perception}}} & Vision & no limitation & 0 \\
        \cline{3-4}
        ~ & ~ & partial limitation & 1 \\
        \cline{3-4}
        ~ & ~ & total limitation & 2 \\
        \cline{2-4}
        ~ & Hearing & no limitation & 0 \\
        \cline{3-4}
        ~ & ~ & partial limitation & 1 \\
        \cline{3-4}
        ~ & ~ & total limitation & 2 \\
    \end{tabular}
    \label{tab:disabilities}
\end{table}

\section{Mock-Up Applications and Examples}
\label{sec:application}
In this section, we present the interactable mock-up for I/O device selection by showing and discussing two examples. This exploration not only highlights the feasibility of our method but also underscores the broader implications and challenges associated with its deployment in application.

\subsection{Graphical User Interface}
\begin{figure*}[p]
    \centering
     \begin{subfigure}[t]{\textwidth}
         \centering
         \includegraphics[width=.9\textwidth]{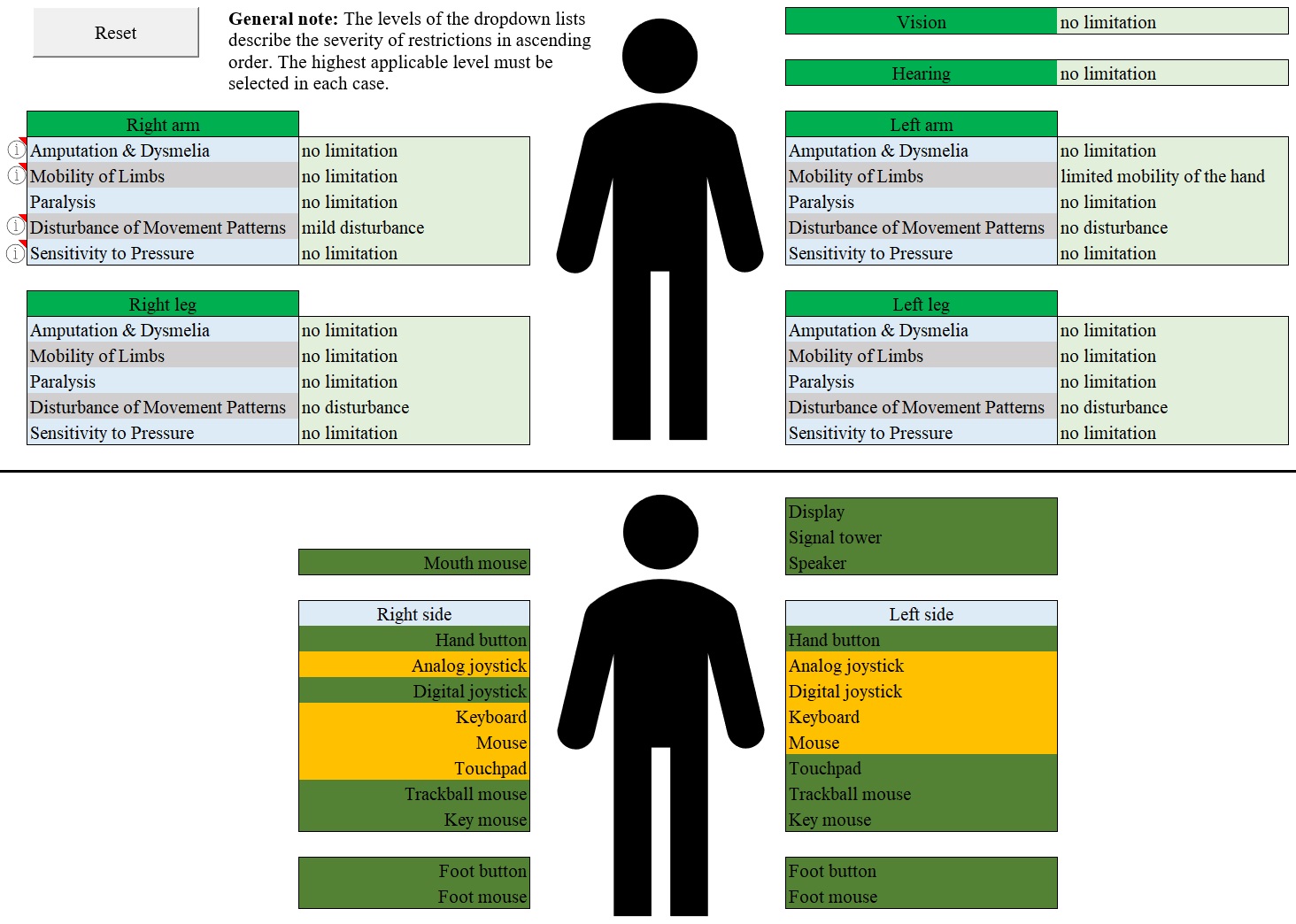}
         \caption{Example 1: A PwD experiencing tremors after a stroke; movement disorder and limited mobility in the left hand.}
         \label{sfig:gui_example1}
     \end{subfigure}
     \hfill
     \begin{subfigure}[t]{\textwidth}
         \centering
         \includegraphics[width=.9\textwidth]{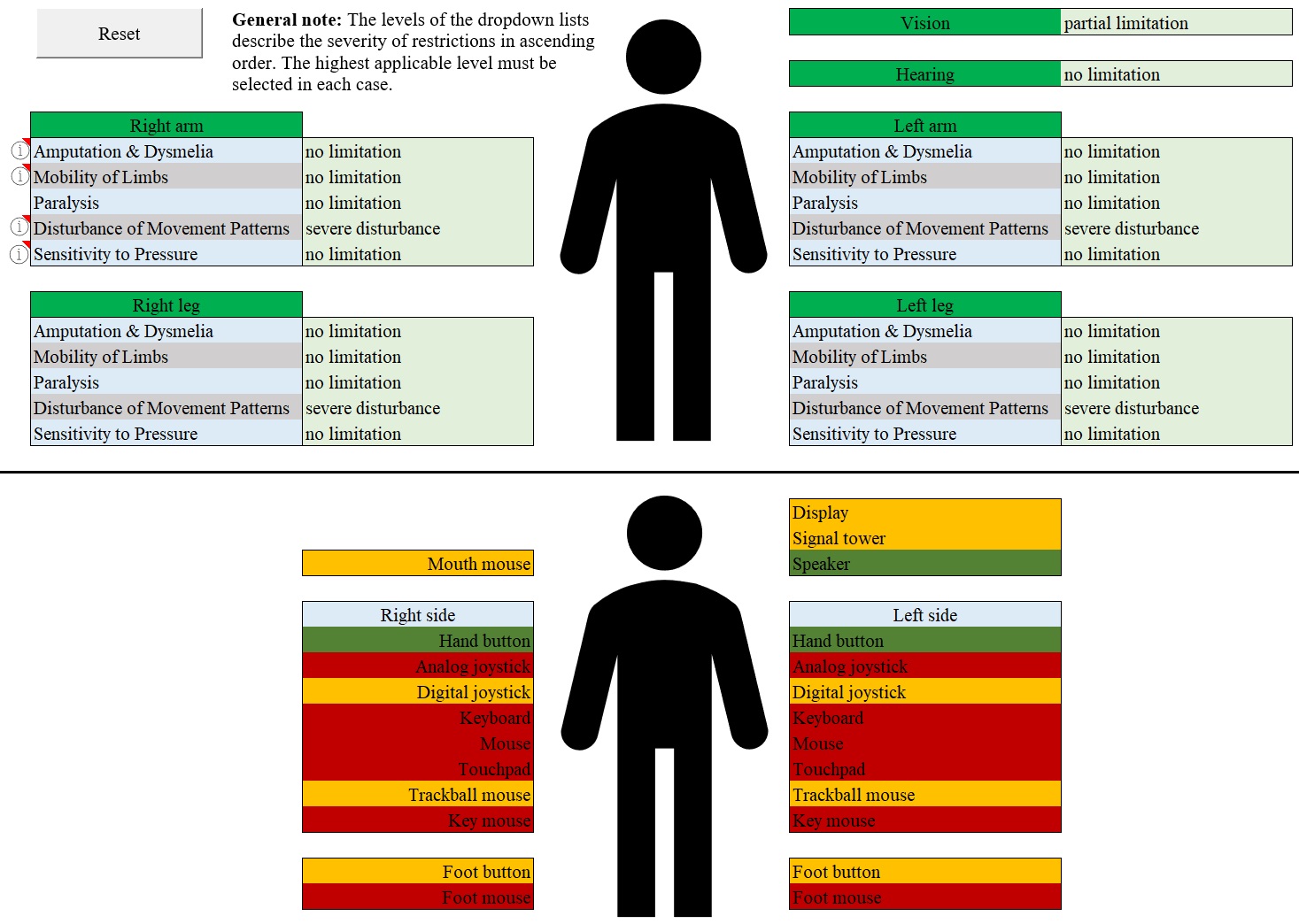}
         \caption{Example 2: A PwD with impaired vision and severely limited mobility caused by Parkinson's disease.}
         \label{sfig:gui_example2}
     \end{subfigure}
    \caption{Mock-up application. Upper view: User sets limitations according to Table~\ref{tab:disabilities}. Lower view: Application outputs device selection. The GUI mock-up is available at~\cite{Zenodo2024}.} 
    \label{fig:gui}
\end{figure*}

The GUI of the mock-up is depicted in Figure~\ref{fig:gui}.
The mock-up is implemented with Microsoft Excel. In the input section in the upper part of Figure~\ref{fig:gui}, users may input specific disabilities for each limb (right and left arm, right and left leg) and relevant cognitive disabilities. Drop-down menus are provided under each category, enabling users to select the appropriate degree of disability from the available options, which are essentially the values of Table~\ref{tab:disabilities}. The output section in the lower part of Figure~\ref{fig:gui} is likewise divided into limbs and perception, displaying usable I/O devices for each. The devices are color-coded to denote the compatibility with the user's disability:
\begin{description}
    \item[Green] indicates that the device can be used by the PwD with the specified disability.
    \item[Yellow] indicates that the PwD's degree of disability is one value higher than the device's requirement. These cases should be validated by the designer, as even categorization in degrees of disability is subject to bias. In addition, for some devices physical or technical compensation may enable operation.
    \item[Red] indicates that the PwD's degree of disability exceeds the device's requirement by more than two values (in one category or as an aggregation of multiple categories), rendering the device unsuitable for operation.
\end{description}
The output is dynamically updated after each input, providing real-time feedback to users. A reset button is provided to clear all inputs and return the interface to its initial state. This simple and user-friendly interface aims to streamline the filtering process and enhance accessibility for non-experts in the design process of human-robot workstations.

\subsection{Application Examples}
The \textbf{first example} examines a PwD experiencing tremors after a stroke, resulting in a movement disorder and limited mobility in the left hand (Figure~\ref{sfig:gui_example1}). The results indicate different devices for the left and right arm due to the difference in disability. It is evi\-dent that analyzing each limb individually is necessary to identify suitable devices. Device prioritization is subjective and dependent on personal preferences. However, it is recommended to prioritize simple devices. In this example, providing the PwD with a mouth joystick is impractical, as the person would need to learn to control the device. As devices known from everyday-life are available (hand buttons, joysticks, touchpads), one of these shall be chosen. The final selection of devices is subject to the designer and needs to be aligned with the preferences of the PwD. Our method rules out non-usable devices to shrink the design space.

The \textbf{second example} covers a PwD with impaired vision and severely limited mobility caused by Parkinson's disease (Figure~\ref{sfig:gui_example2}). It is important to note the extra effort required to configure the inputs, as the movement disorder must be entered for each limb separately even though all are equally limited. The method only recommends buttons as usable devices without further examination. However, controlling the underlying process with buttons is not always practical or may require a large number of buttons or complex sequences of button presses. Due to partial limitation of vision, monitors and signal towers are categorized as yellow. In application, the usability is dependent on the size and placing of the devices, and the severity of the partial limitation of the PwD. However, the signal tower is more likely to be used than the display, as the representation of information in form of colored signals is simpler than reading from a monitor. In addition, perceiving just the presence or absence of colored light is available in case of more sever vision impairment. Note that latter requires that the person can perceive color.

Comparing the examples, it becomes apparent that buttons are most likely to be selected. This is due to their simplicity in operation. Further, we were able to show that the categories and degrees of disabilities indicate feasible devices, even in PwD with multiple disabilities. Here lies the strength of our method, as it allows the designer to better keep track of the implications of the various disabilities and their severity. The degrees of disability are chosen detailed enough to allow a nuanced assessment, also in case of co-dependency. However, before the mock-up may be used in application, some quality of life features need to be added, e.g., an agglomeration of disabilities affecting both sides and/or multiple limbs equally. Further, the method lacks prioritization. This would require to analyze way more then devices and the PwD's degree of disabilities. To also craft a prioritized list of devices, the full process, environment, and ergonomic standards would need to be consulted. This would raise the complexity of the method and potentially turn it into an expert only tool. Instead, we propose to first use our method to shrink the design space and then consult further aspects of the work environment and find a suitable device in close contact with the individual PwD.

\section{Conclusions}
In this work, we introduced a method for selecting suitable I/O devices that are necessary for the underlying process and usable for PwD. The method aligns with VDI~2221, allowing for easy integration into existing product development processes of human-robot workstations. The method selects feasible devices, allows for nuanced assessment, and works with multiple and complex disabilities. To achieve this, we first analyzed the basic structure of human-robot workstations and their subsystems. We identified the necessary devices for a work process and derived specific devices that are suitable for PwD. These devices were then matched with the relevant disabilities using a requirement matrix, which indicates the maximum level of disability of a PwD to be able to operate a device. Finally, we implemented our method in a mock-up application with a GUI and provided two application examples. These two examples demonstrated the feasibility and simplicity of our method. By specifying an individual's disabilities through drop-down lists, the mock-up outputs suitable devices. This reduces the design space and enables tracking of the implications of various disabilities, including their severity. The proposed method currently lacks prioritization and considerations for the implementation of the workstation. Therefore, when selecting devices, the designer must consider factors such as the preferences of people with disabilities, device positioning, and adherence to the two-senses principle. The method does not replace the designer but is aimed to support in decision making during the design process of human-robot workstations.

\section*{Acknowledgement}
We like to thank Radite Adyanawa, Carla Angerhausen, and Jonas Braun who supported in developing the methodology and implementing the GUI mock-up.

\balance


\bibliographystyle{IEEEtranUrldate}
\bibliography{smc2024_refs.bib}

\end{document}